\documentclass[runningheads]{llncs}

\usepackage[T1]{fontenc}

\usepackage{graphicx,verbatim}
\usepackage{booktabs}
\usepackage{tabularx}
\usepackage{multirow}
\usepackage[flushleft]{threeparttable}
\usepackage{array, makecell}
\usepackage{multirow}
\usepackage{amsmath}
\usepackage{mathtools}
\usepackage{url}
\usepackage{layouts}
\usepackage[dvipsnames]{xcolor}
\usepackage[symbol]{footmisc}
\usepackage[hidelinks]{hyperref}%

\DeclareMathOperator*{\argmin}{arg\,min}

\begin{document}

\title{Learning Disease State from Noisy Ordinal Disease Progression Labels}

\author{
Gustav Schmidt\inst{1} \and
Holger Heidrich\inst{2} \and
Philipp Berens\inst{1} \and
Sarah M\"uller\inst{1}
}

\institute{
    Hertie Institute for AI in Brain Health, University of T\"ubingen, Germany \and 
    Department of Computer Science, University of T\"ubingen, Germany \\
\email{sar.mueller@uni-tuebingen.de}}

\authorrunning{G. Schmidt et al.}

\maketitle           

\begin{abstract}
Learning from noisy ordinal labels is a key challenge in medical imaging. In this work, we ask whether ordinal disease progression labels (\textit{better}, \textit{worse}, or \textit{stable}) can be used to learn a representation allowing to classify disease state. For neovascular age-related macular degeneration (nAMD), we cast the problem of modeling disease progression between medical visits as a classification task with ordinal ranks. To enhance generalization, we tailor our model to the problem setting by (1) independent image encoding, (2) antisymmetric logit space equivariance, and (3) ordinal scale awareness. In addition, we address label noise by learning an uncertainty estimate for loss re-weighting. Our approach learns an interpretable disease representation enabling strong few-shot performance for the related task of nAMD activity classification from single images, despite being trained only on image pairs with ordinal disease progression labels\footnote[1]{\url{https://github.com/berenslab/Learning-Disease-State}}.


\end{abstract}

\section{Introduction}
Changes apparent in medical images can be informative about the progression of a disease, playing a critical role in guiding clinical decision making, particularly for conditions requiring timely interventions. One such example is the treatment of neovascular age-related macular degeneration (nAMD) with anti-vascular endothelial growth factor (anti-VEGF) therapy. Here, the treatment is guided by the presence and extent of exudative signs, such as intraretinal and subretinal fluid as relevant biomarkers \cite{Schmidt2016paradigm}. These are best assessed with optical coherence tomography (OCT) imaging. Accurate prediction of disease progression in this context could help to optimize treatment schedules and improve patient outcomes. Different deep learning approaches have been proposed to analyze AMD based on OCT B-scans (i.e. individual images), including disease activity classification \cite{ayhan2023,Koseoglu2023DeepAMD}, biomarker identification \cite{holland2024,hanson2023optical}, and disease progression modeling \cite{Rivail2019,Emre2022}. 

\begin{figure}
    \centering
    \includegraphics[width=\linewidth]{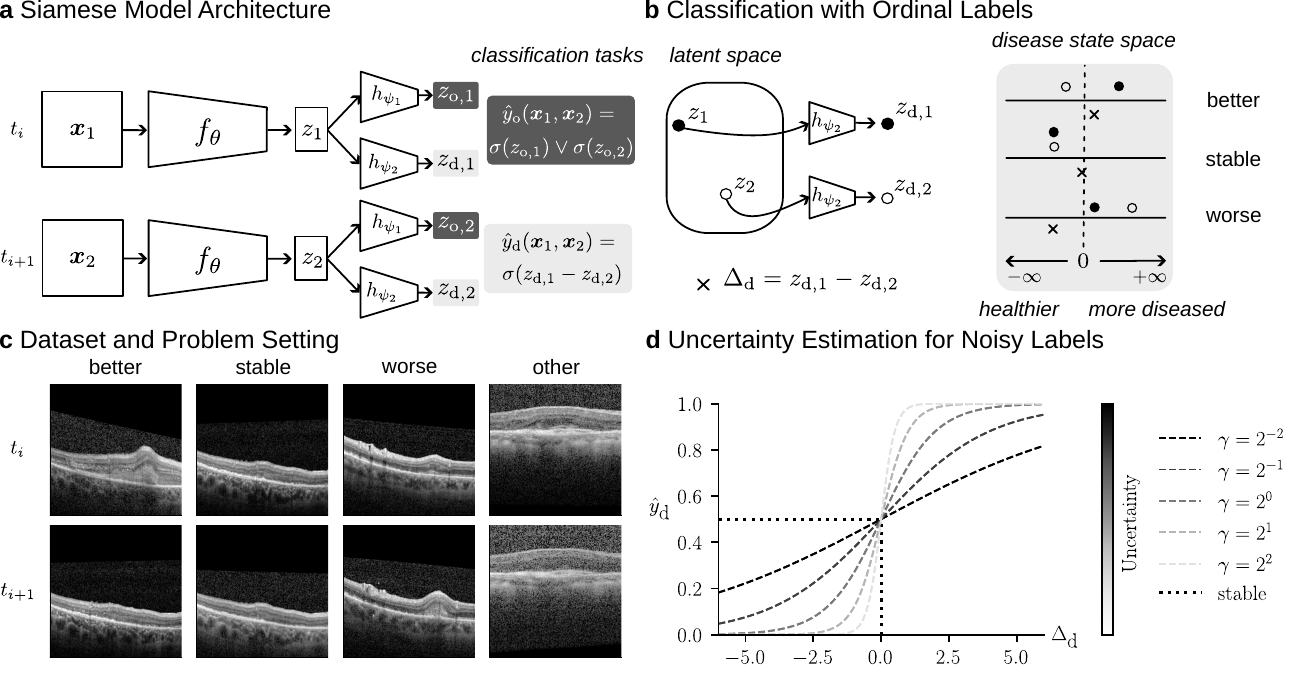}
    \caption{Overview figure of the proposed approach. (a) Our Siamese model analyses two OCT B-scans (c), outputs the probability of being ungradable ($\hat{y}_\textrm{o}$), and predicts the disease progression ($\hat{y}_\textrm{d}$). (b) For disease progression prediction our model internally learns a disease state space ($z_\textrm{d}$) for each image. (d) Moreover, we fit an uncertainty estimate for each image pair to account for label noise. $\sigma(\cdot)$ is the sigmoid function.}
    \label{fig:overview-figure}
\end{figure}

Interestingly, judging disease progression between OCT B-Scan pairs is generally believed to be easier and less biased for clinicians than assigning categorical severity scores to individual B-Scans used in the studies cited above, as it is closer to the clinical task performed in a routine assessment. Here, we ask whether we can use such coarse, ordinal information about whether nAMD has improved, worsened, or remained stable between two visits (Fig.\,\ref{fig:overview-figure}\,\textbf{c}) to learn about the underlying disease state. While ordinal regression has been explored in machine learning \cite{coral2020,corn2023,garg2020} and medical imaging \cite{Baek2024,Tang2023,Zhuangzhi2025,Polat2025,Hoebel2023}, learning from noisy ordinal disease progression labels about the underlying disease state remains unexplored.

To this end, we used the MARIO challenge dataset \cite{mario} from MICCAI 2024, which consists of matched, labeled image pairs indicating the relative disease state or disease progression between two patient visits. We frame disease progression between medical visits as a classification task with ordinal ranks (Fig.\,\ref{fig:overview-figure}\,\textbf{a,b}), akin to ordinal regression. To enhance generalization and interpretability, we tailored our model to the problem setting using four key steps: 
\begin{enumerate}
    \item \textbf{Independent image encoding}: We encode disease-related features for each image independently (Fig.\,\ref{fig:overview-figure}\,\textbf{a}) to ensure that even though the model is trained only on labeled image pairs, it retains meaningful individual image representations for disease state (Fig.\,\ref{fig:overview-figure}\,\textbf{b}). 
    \item \textbf{Antisymmetric logit space equivariance}: Since disease progression labels primarily capture differences between image pairs, we directly model these differences in the logit space (Fig.\,\ref{fig:overview-figure}\,\textbf{b}). By representing the differences along a continuous scale, we are able to better capture the continuous nature of disease progression.
    \item \textbf{Ordinal scale awareness}: Unlike conventional ordinal regression, our method accounts not only for the relative ordering of ordinal variables but also for the known symmetric distances between labels, leading to a more structured and interpretable representation of disease progression. 
    \item By \textbf{uncertainty-aware loss re-weighting} \cite{label_noise_2024} we mitigate the impact of label noise (Fig.\,\ref{fig:overview-figure}\,\textbf{d}). This accounts for different sources of label noise, allowing the model to better capture inherent ambiguities in clinical grading. 
\end{enumerate}

We then show that the learned representation leads to strong few-shot out-of-distribution performance on an in-house OCT dataset labeled for nAMD activity.

\section{Methods}

\subsection{Dataset and Preprocessing}
For training and evaluating our models, we worked with the MARIO challenge data \cite{mario}. We used the development data consisting of 14,496 labeled OCT B-scan pairs from 68 patients, with up to 10 visits per patient. All OCT volumes were acquired with Heidelberg Spectralis and volumes were registered between consecutive visits using the Spectralis follow-up option. To standardize the input dimension, we padded all images to match the largest resolution occurring in the dataset ($496 \times 1024$). Furthermore, we applied training-time data augmentation, ensuring that the same augmentations were applied for both images out of a pair. Augmentations included random resize cropping between 20\%-100\% of the original size, resized to $224 \times 385$, and random horizontal flipping. We split our dataset patient-wise in 85\% for training (5-fold cross-validation) and 15\% for testing.

\subsection{Problem Setting}\label{subsec:problem}
The B-scan image pairs were annotated by ophthalmologists into seven initial classes, which were  later simplified into three disease progression classes \textit{better}, \textit{worse}, \textit{stable}, and one \textit{other} category for ungradable image pairs (Fig.\,\ref{fig:overview-figure}\,\textbf{c}). Therefore, the challenge framed the task as a 4-class classification problem, where the disease progression labels are on an ordinal scale with symmetric distances -- the \textit{stable} class is between \textit{better} and \textit{worse}. 

\subsection{Obtaining Disease State from Coarse Progression Labels}
We used a Siamese neural network to process labeled B-scan pairs (Fig.\,\ref{fig:overview-figure}\,\textbf{a}). However, instead of na\"ively training the Siamese network with a categorical cross-entropy loss for the 4-class problem, we tailored our model to the problem setting. Because the \textit{other} class is a very different category than the disease progression categories, we separated disease information and the \textit{other} class with two independent heads ($\boldsymbol{z}_\textrm{d}$ and $\boldsymbol{z}_\textrm{o}$ in Fig.\,\ref{fig:overview-figure}\,\textbf{b}) and treated both heads as independent classification tasks. This has the advantage that our model predicts disease embeddings $z_\textrm{d}$ for each B-scan individually, even though we only train it on labeled B-scan pairs. Then, we reasoned that disease progression prediction was mainly about the difference between the image pairs. In theory, we could subtract registered B-scan pairs from each other and operate on difference images, however, to be more robust to possible registration errors and other noise differences between visits, we decided to compute the differences in the logit (unnormalized log probability) space. Therefore, we predicted the disease progression with
\begin{equation}
    \hat{y}_\textrm{d}(\boldsymbol{x}_1,\boldsymbol{x}_2) = \sigma(\Delta_\textrm{d}) = \frac{1}{1+e^{-\Delta_\textrm{d}}},\, \Delta_\textrm{d} = z_{\textrm{d},1} - z_{\textrm{d},2} .
\end{equation}
Estimating disease progression is akin to a binary classification task with the two classes $\textit{worse}\mathrel{\hat=}0$ and $\textit{better}\mathrel{\hat=}1$. However, our setup extended binary classification with a third label $\textit{stable}\mathrel{\hat=}0.5$. By setting the labels in this way, they inherently follow the order of our ordinal labels by positioning the \textit{stable} class between \textit{better} and \textit{worse} (Fig.\,\,\ref{fig:overview-figure}\,\textbf{d}). Moreover, regarding the true continuous nature of disease progression behind discrete labels, our approach allows the model to position the \textit{better} and \textit{worse} predictions on a continuous scale in logit space. This extension also leads to an antisymmetric equivariance with respect to the image order, so that $\hat{y}_\textrm{d}(\boldsymbol{x}_1,\boldsymbol{x}_2) = 1- \hat{y}_\textrm{d}(\boldsymbol{x}_2,\boldsymbol{x}_1)$. This assumes that matching pairs in both time directions (forward and backward) helps learning about disease progression. 

For the \textit{other} binary classification task, we merged the two \textit{other} head predictions for the individual images with a logical OR operation -- if at least one image of the pair is ungradable, e.g., due to noise, then the model should already predict the \textit{other} class. We applied De Morgan's laws to reformulate the \textit{other} prediction to the differential computation
\begin{align}
    \hat{y}_\textrm{o}(\boldsymbol{x}_1, \boldsymbol{x}_2) &= p((\boldsymbol{x}_1 \text{ is \textit{other}}) \lor (\boldsymbol{x}_2 \text{ is \textit{other}})) \\
    &= 1 - (1 - \sigma(z_{\textrm{o},1})) \cdot (1- \sigma(z_{\textrm{o},2})) = 1 - \sigma(-z_{\textrm{o},1}) \cdot \sigma(-z_{\textrm{o},2})
\end{align}
where $\sigma(x)$ is the sigmoid function with the property $1-\sigma(x)=\sigma(-x)$. \\

\subsection{Modeling Label Noise}
To model label noise and account for uncertain examples, 
we included a learnable slope parameter $\gamma$ (Fig.\,\ref{fig:overview-figure}\,\textbf{d}) into the disease progression tasks sigmoid function
\begin{align}
    \hat{y}_\textrm{d}(\boldsymbol{x}_1, \boldsymbol{x}_2) &= \sigma(\Delta_\textrm{d}, \gamma) = \frac{1}{1+e^{-\gamma\cdot \Delta_\textrm{d}}}
\end{align}
to give our model the possibility to set this parameter for every B-scan pair at training time.
Intuitively, we interpreted the $\gamma$ values as an uncertainty estimate for each image pair, where lower and higher values than $\gamma=1$ refered to higher and lower uncertainty, respectively (Fig.\,\ref{fig:overview-figure}\,\textbf{d}). However, $\gamma$ could also be misused as a shortcut by the model for ``hard'' to classify image pairs by setting $\gamma$ very low and hence achieve lower loss on those samples. Therefore, we introduced a regularizer, which regularizes $\gamma$ to be close to its default value $\gamma=1$. We defined $\gamma = 2^\alpha$ and add $|\alpha|$ as a regularizer to the final optimization problem
\begin{align}
    \theta^*, \psi_1^*, \psi_2^*, \alpha^* = \argmin_{\theta, \psi_1, \psi_2, \alpha} \text{BCE}(y_\textrm{o}, \hat{y}_\textrm{o}) + \text{BCE}(y_\textrm{d}, \hat{y}_\textrm{d}) + \lambda |\alpha|
\end{align}
with $\text{BCE}$ as the binary cross entropy loss.
We balanced the dataset with a weighted random sampler and trained all models with 5-fold cross-validation, a ResNet50 \cite{he2015deepresiduallearningimage} backbone and AdamW \cite{loshchilov2019decoupledweightdecayregularization} optimizer ($\text{lr} = 10^{-4}$). All models were trained for 60 epochs and we selected the best model for evaluation by the best validation loss. For our models with noise estimation, we set $\lambda=0.15$, selected by grid search.

\section{Results}

\subsection{Disease Progression Classification}
\begin{table}[b]
    \centering
     \caption{Classification performance (in \%) for disease progression.}
    \begin{tabularx}{\textwidth}{l@{\hskip 0.15in} c@{\hskip 0.05in} c@{\hskip 0.05in} c@{\hskip 0.05in} c@{\hskip 0.05in} c@{\hskip 0.05in} c@{\hskip 0.05in}} 
        \toprule
        \textbf{Model} & \textbf{F1 Score} & \textbf{Rk-corr.} & \textbf{Specificity} & \textbf{Bal. Acc.} & \textbf{Precision} & \textbf{Recall} \\
        \midrule
        na\"ive clf. & $70\pm 5$ & $44 \pm 4$ & $87 \pm 1$ & $60 \pm 3$ & $57 \pm 3$ & $60 \pm 3$ \\
        ours & $61 \pm 7$ & $36 \pm 5$ & $86 \pm 1$ & $59 \pm 5$ & $47 \pm 3$ & $59 \pm 5$ \\
        ours + noise estim. & $60 \pm 7$ & $36 \pm 6$ & $86 \pm 2$ & $60 \pm 4$ & $47 \pm 4$ & $60 \pm 4$ \\
        \bottomrule
    \end{tabularx}
    \label{tab:classification_performance}
\end{table}

We used the MARIO OCT B-scan dataset \cite{mario} to train our architecture suitable for handling ordinal disease progression labels (Fig.\,\ref{fig:overview-figure}) and first evaluated it for disease progression classification on the metrics of the MARIO challenge \cite{mario-proceedings}. We compared its performance to a na\"ive classifier (clf.) trained for a 4-class problem with categorical cross-entropy. For our model, to assign each point in $\hat{y}_\textrm{d}$ to a class, we optimized a symmetric decision boundary around $0.5$ based on the validation data. Our model performed comparable to the the na\"ive classifier (Table\,\ref{tab:classification_performance}) for specificity, balanced accuracy and recall. However, in other metrics like the F1 score, our model showed reduced performance, mainly due to the performance for the \textit{stable} class. While the focus of this work was not on optimizing performance for the disease progression classification task, our model would have been placed 17/21 on the MARIO leaderboard, when evaluated on 15\% of the training data in the cross-validation setting (as we did not have access to the MARIO validation set which was used to rank the participants).
\begin{figure}[t!]
    \centering
    \includegraphics[width=\linewidth]{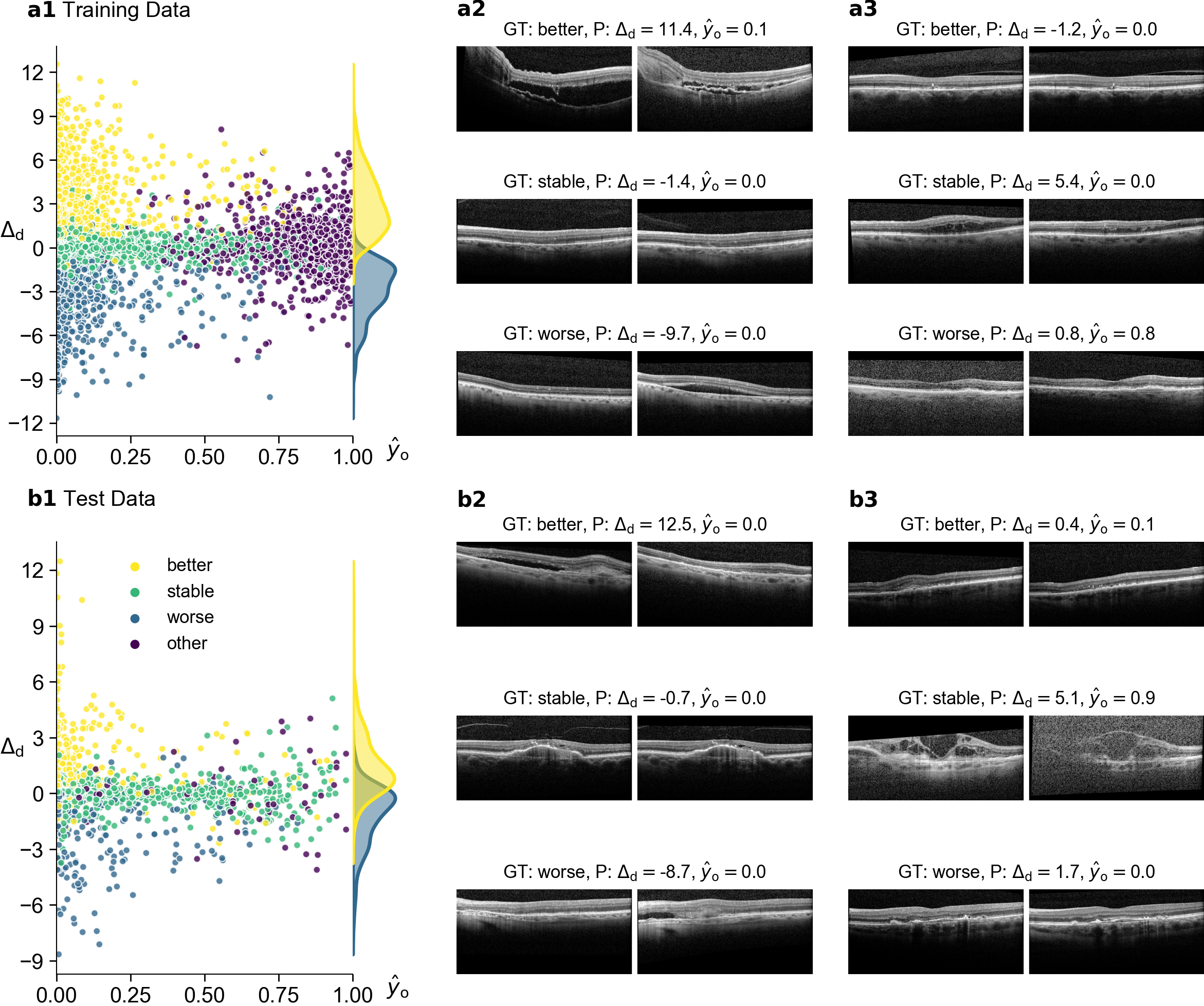}
    \caption{Our model internally maps coarse ordinal labels onto a continuous scale that preserves ordinal ranks. (a1) Continuous disease progression variable $\Delta_\textrm{d}$ vs. the probability of belonging to the \textit{other} class for training data. (a2) Examples with low probability of \textit{other} illustrating the graded nature of disease progression judgments, with the magnitude of $\Delta_\textrm{d}$ corresponding to the magnitude of observed changes. (a3) Examples with ''incorrect`` predictions compared to the ground truth labels. Top: Labeled as \textit{better}, but looks stable, as indicated by $\Delta_\textrm{d}$. Middle: Labeled as \textit{stable}, but looks improved, reflected in $\Delta_\textrm{d}$. Bottom: Labeled \textit{worse}, but looks stable and noisy captured by $\hat{y}_\textrm{o}$. (b1-b3) as in (a1-a3) but for test data. (GT: ground truth, P: prediction).}
    \label{fig:continuous_disease_progression}
\end{figure}

However, compared to the na\"ive classifier, our model internally learned a continuous disease progression representation $\Delta_\textrm{d}$. We qualitatively  analyzed what additional information this representation could provide both for the training and test data (Fig.\,\ref{fig:continuous_disease_progression}\,\textbf{a1},\textbf{b1}). When selecting image pairs with large values in $\Delta_\textrm{d}$, large changes between the images were visible, indicating that the representation accurately reflected disease progression information (Fig.\,\ref{fig:continuous_disease_progression}\,\textbf{a2},\textbf{b2}). Additionally, we probed the representation to understand why some image pairs were incorrectly predicted or had low confidence predictions (Fig.\,\ref{fig:continuous_disease_progression}\,\textbf{a3},\textbf{b3}). We found that in many cases, the ground truth labels did not match well what was visible on the images, as the images did not show clear evidence for the labeled class but rather for one of the others, indicating that many ground truth labels were noisy. Also, the \textit{other} class prediction helped to detect corrupted images with high success.

\subsection{Learning an Uncertainty Parameter for Label Noise}
\begin{figure}[t!]
    \centering
    \includegraphics[width=\textwidth]{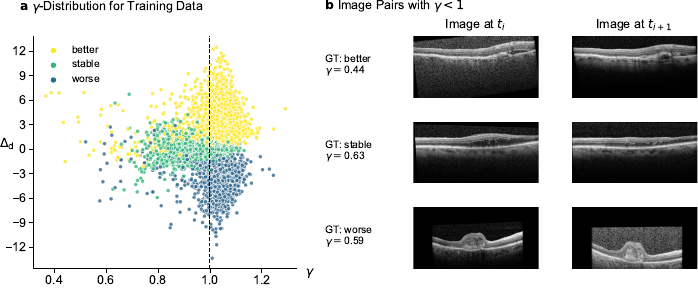}
    \caption{Modeling label noise through uncertainty parameter. (a) The uncertainty parameter $\gamma$ is lower for images leading to $\Delta_\textrm{d}\approx 0$, indicating higher uncertainty. (b) Example images show that $\gamma$ captures different sources of noise, including acquisition noise, mislabeling and failed registration (top-bottom). (GT: ground truth).}
    \label{fig:gamma_model}
\end{figure}

Our observation that ground truth labels were unreliable and noisy in many cases was corroborated by the fact that adjacent B-scans in OCT volumes sometimes received different labels, despite showing similar structural patterns. Therefore, we extended our model with a mechanism to discount noisy labels during training, learning an uncertainty parameter $\gamma$ for each B-scan pair (Fig.\,\ref{fig:gamma_model}\,\textbf{a}). In fact, the learned $\gamma$ was smaller for image pairs close to the decision boundaries. We found that image pairs with low $\gamma$ values corresponded to noise from acquisition, mislabeling or failed registration (Fig.\,\ref{fig:gamma_model}\,\textbf{b}, top to bottom). To verify this qualitative finding, we analyzed the $\gamma$ values of examples adjacent in the OCT volume but labeled differently. We found that 24\% of examples labeled \textit{better} next to \textit{worse} had $\gamma$ values below 0.85, compared to 15\% for both \textit{better} to \textit{stable} and \textit{worse} to \textit{stable} transitions, while only around 8\% for cases remaining in the same state (e.g. \textit{better} to \textit{better}).

\subsection{Out-of-Distribution Few-Shot Disease Activity Classification}

\begin{figure}[t!]
    \centering
    \includegraphics[width=\linewidth]{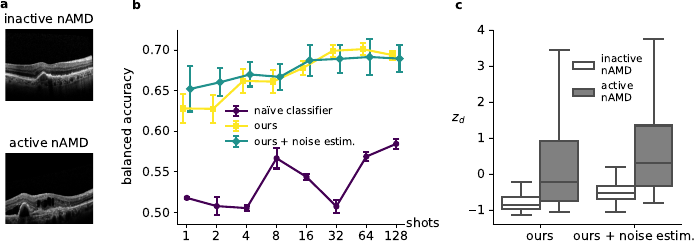}
    \caption{Our model allows accurate out-of-distribution few-shot nAMD activity classification. (a) Example OCT B-scans from the in-house dataset. (b) Balanced accuracy as a function of the number of single B-scans per class for our method (yellow), our method with noise estimation (green) and a na\"ive classifier (purple). (c) Inactive vs. active nAMD B-scans lead to distributions in the $z_d$ space (here shown for one fold).}
    \label{fig:few_shot_ukt}
\end{figure}

Because our model was able to extract a meaningful continuous disease progression representation from the coarse ordinal labels, we hypothesized that the representation learned by $z_d$ could be used for performing a related task on an out-of-distribution dataset. To this end, we mapped 1,998 B-scans from the internal OCT dataset acquired with Heidelberg Spectralis (Fig.\,\ref{fig:few_shot_ukt}\,\textbf{a}) with binary nAMD activity labels (\textit{inactive} vs. \textit{active}) to the $z_d$ space. The in-house dataset is not publicly available but was first used in \cite{ayhan2023} and approved by the local institutional ethics committee. We then optimized the decision boundary between inactive and active nAMD patients for the $z_d$ logits on a small set of images (we call $k$-shots, where $k$ is the number of B-scans per class) without retuning the representation in any way. Compared to a logistic regression classifier on the embeddings of the na\"ive model, our models showed better nAMD activity classification performance, even when the decision boundary was determined using very little data (Fig.\,\ref{fig:few_shot_ukt}\,\textbf{b}). Furthermore, the model with noise estimation proved to be the most robust model in terms of balanced accuracy, especially for very few shots (Fig.\,\ref{fig:few_shot_ukt}\,\textbf{b},\textbf{c}).

\section{Discussion}
We introduced a novel approach to learn from coarse ordinal disease progression labels (\textit{better}, \textit{stable} or \textit{worse}) by tailoring a deep learning model to the task, showing strong few-shot generalization performance on an in-house out-of-distribution OCT dataset on a related, but not identical task of nAMD activity classification. Our model provided enhanced interpretability by learning a continuous disease progression space internally, compared to models directly targeting the classification task. Additionally, by explicitly modeling the noise in the ordinal labels, we were able to  explore various sources of label noise and show more robust out-of-distribution performance.

In future work, an enhanced backbone architecture could be used to improve disease progression classification on the MARIO challenge data. Furthermore, we observed that the \textit{stable} class and our noise model appear to interfere, leading the model to assign lower gamma values to this class. Therefore, exploring alternative approaches for modeling label noise (e.g., \cite{englesson2024robust}) could offer a promising direction for future work. Finally, the disease state space currently learned by the model is restricted to one-dimension -- an insightful ablation experiment would be to investigate how the performance would change if we allow a higher-dimensional latent representation.

\newpage

\begin{credits}
\subsubsection{\ackname} We thank Sebastian Damrich, Verena Jasmin Hallitschke, and Julius Gervelmeyer for helpful discussions and their valuable feedback. This project was supported by the Hertie Foundation and by the Deutsche Forschungsgemeinschaft under Germany’s Excellence Strategy with the Excellence Cluster 2064 ``Machine Learning — New Perspectives for Science'', project number 390727645. PB is a member of the Else Kr\"oner Medical Scientist Kolleg ``ClinbrAIn: Artificial Intelligence for Clinical Brain Research''. The authors thank the International Max Planck Research School for Intelligent Systems (IMPRS-IS) for supporting SM.

\subsubsection{\discintname}
The authors have no competing interests to declare that are relevant to the content of this article.
\end{credits}

\bibliographystyle{splncs04}
\bibliography{references.bib}

\end{document}